\def\year{2020}\relax
\documentclass[letterpaper]{article} 
\usepackage{aaai20}  
\usepackage{times}  
\usepackage{helvet} 
\usepackage{courier}  
\usepackage[hyphens]{url}  
\usepackage{graphicx} 
\usepackage{graphicx}  
\usepackage{amsmath}
\usepackage{amsfonts}
\usepackage{multirow}
\usepackage{tikz}
\usepackage{comment}
\usepackage{multicol}
\usepackage{multirow}
\usepackage{booktabs}
\usepackage{pgfplots}
\usetikzlibrary{shapes, positioning, decorations.pathreplacing, shapes.multipart,arrows,chains,calc,plotmarks}
\usepackage{url}
\usepackage{subfig}

\newenvironment{itemize*}%
  {\begin{itemize}%
    \setlength{\itemsep}{3pt}%
    \setlength{\parskip}{3pt}}%
  {\end{itemize}}

    \newenvironment{enumerate*}%
  {\begin{enumerate}%
    \setlength{\itemsep}{3pt}%
    \setlength{\parskip}{3pt}}%
  {\end{enumerate}}

\def\bH{\mathbf{H}}

\nocopyright
\newcommand{\citet}[1]{\citeauthor{#1}~\shortcite{#1}}

\title{Multi-Scale Self-Attention for Text Classification}
\author{Qipeng Guo\footnotemark[3]\thanks{Work done during internship at AWS Shanghai AI Lab.}, Xipeng Qiu\footnotemark[3] \thanks{Corresponding author: Xipeng Qiu (xpqiu@fudan.edu.cn).}, Pengfei Liu\footnotemark[3], Xiangyang Xue\footnotemark[3], Zheng Zhang\footnotemark[4]\\
\footnotemark[3] Shanghai Key Laboratory of Intelligent Information Processing, Fudan University\\
\footnotemark[3] School of Computer Science, Fudan University\\
\footnotemark[4] AWS Shanghai AI Lab \\
\footnotemark[4] New York University Shanghai \\
\{qpguo16, xpqiu, pfliu14, xyxue\}@fudan.edu.cn, zz@nyu.edu
}

\begin{document}

\maketitle

\begin{abstract}
In this paper, we introduce the prior knowledge, multi-scale structure, into self-attention modules.  We propose a Multi-Scale Transformer which uses multi-scale multi-head self-attention to capture features from different scales. Based on the linguistic perspective and the analysis of pre-trained Transformer (BERT) on a huge corpus, we further design a strategy to control the scale distribution for each layer. Results of three different kinds of tasks (21 datasets) show our Multi-Scale Transformer outperforms the standard Transformer consistently and significantly on small and moderate size datasets.
\end{abstract}

\section{Introduction}

Self-Attention mechanism is widely used in text classification tasks, and models based on self-attention mechanism like Transformer \cite{DBLP:conf/nips/VaswaniSPUJGKP17}, BERT \cite{DBLP:journals/corr/abs-1810-04805} achieves many exciting results on natural language processing (NLP) tasks, such as machine translation, language modeling \cite{DBLP:journals/corr/abs-1901-02860}, and text classification. Recently, \citet{radford2018improving} points out the weakness of self-attention modules, especially the poor performance on small and moderate size datasets. Although the pre-training on huge corpus could help this, we believe the fundamental reason is that self-attention module lacks suitable inductive bias, so the learning process heavily depends on the training data. It is hard to learn a model with good generalization ability from scratch on a limited training set without a good inductive bias.

Multi-Scale structures are widely used in computer vision (CV), NLP, and signal processing domains. It can help the model to capture patterns at different scales and extract robust features.
Specific to NLP domain, a common way to implement multi-scale is the hierarchical structure, such as convolutional neural networks (CNN) \cite{kalchbrenner2014convolutional},
multi-scale recurrent neural networks (RNN) \cite{chung2016hierarchical}
tree-structured neural networks \cite{DBLP:conf/emnlp/SocherPWCMNP13,DBLP:conf/acl/TaiSM15} and hierarchical attention \cite{yang2016hierarchical}.
The principle behind these models is the characteristic of language: the high-level feature is the composition of low-level terms.
With hierarchical structures, these models capture the local compositions at lower layers and non-local composition at high layers. This division of labor makes the model less data-hungry.

However, for self-attention modules, there is no restriction of composition bias. The dependencies between words are purely data-driven without any prior, leading to easily overfit on small or moderate size datasets.

In this paper, we propose a multi-scale multi-head self-attention (MSMSA), in which each attention head has a variable scale. The scale of a head restricts the working area of self-attention. Intuitively, a large scale makes the feature involving more contextual information and being more smooth. A small scale insists on the local bias and encourages the features to be outstanding and sharp.
Based on MSMSA, we further propose multi-scale Transformer, consisting of multiple MSMSA layers.
Different from the multi-scale in hierarchical structures, each layer of multi-scale Transformer consists of several attention heads with multiple scales, which brings an ability to capture the multi-scale features in a single layer.

Contributions of this paper are:

\begin{itemize*}
    \item We introduce the multi-scale structure into self-attention framework, the proposed model Multi-Scale Transformer can extract features from different scales.
    \item Inspired by the hierarchical structure of language, we further develop a simple strategy to control the scale distribution for different layers. Based on the empirical result on real tasks and the analysis from BERT, we suggest using more small-scale attention heads in shallow layers and a balanced choice for different scales in deep layers.
    \item The building block of Multi-Scale Transformer, multi-scale multi-head self-attention provides a flexible way to introduce scale bias (local or global), and it is a replacement of the multi-head self-attention and position-wise feed-forward networks.
    \item Results on three tasks (21 datasets) show our Multi-Scale Transformer outperforms the standard Transformer consistently and significantly on small and moderate size datasets.
\end{itemize*}

\section{Background}
Self-attention and its extend architecture, Transformer, achieved many good results on NLP tasks.
Instead of utilizing CNN or RNN unit to model the interaction between different words, Transformer achieves pair-wised interaction by attention mechanism.
\paragraph{Mulit-Head Self-attention}
The main component of the Transformer is the multi-head dot-product attention, which could be formalized as follows.
Given a sequence of vectors $\mathbf{H} \in \mathbb{R}^{N \times D}$, where $N$ is the length of the sequence and the $D$ is the dimension of the vector. When doing multi-head self-attention, the module projects the $\mathbf{H}$ into three matrices: the query $\mathbf{Q}$, the key $\mathbf{K}$ and the value $\mathbf{V}$. These three matrices would be further decomposed into $N'$ sub-spaces which corresponds to the $N'$ heads and each head has $D'$ units.
\begin{align}
    & \mathrm{MSA}(\mathbf{H}) = [\text{head}_1,\cdots, \text{head}_{N'}] \mathbf{W}^O, \\
    & \text{head}_i = \text{softmax}(\frac{\mathbf{Q}_i \mathbf{K}_i^T}{\sqrt{D'}})\mathbf{V}_i, \\
    & \mathbf{Q} = \mathbf{H} \mathbf{W}^Q, \mathbf{K} = \mathbf{H} \mathbf{W}^K, \mathbf{V} = \mathbf{H} \mathbf{W}^V,
\end{align}
where $\mathrm{MSA}(\cdot)$ represents the Multi-head Self-Attention, and $\mathbf{W}^Q$, $\mathbf{W}^K$, $\mathbf{W}^V$, $\mathbf{W}^O$ are learnable parameters.

\paragraph{Transformer}
Each layer in the Transformer consists of a multi-head self-attention and a FFN layer (also called Position-wise Feed-Forward Networks in \citet{DBLP:conf/nips/VaswaniSPUJGKP17}). The hidden state of $l+1$ layer, $\mathbf{H}_{l+1}$ could be calculated as followings.
\begin{align}
    \mathbf{Z}_l =& \mathrm{norm}(\mathbf{H}_l + \text{MSA}(\mathbf{H}_l)), \\
    \mathbf{H}_{l+1} =& \mathrm{norm}(\mathbf{Z}_l + \text{FFN}(\mathbf{Z}_l)),\label{eq:trans-fnn}
\end{align}
where $\mathrm{norm}(\cdot)$ means the layer normalization \cite{DBLP:journals/corr/BaKH16}. In addition, the Transformer augments the input features by adding a positional embedding since the self-attention could not capture the positional information by itself.

Despite its effectiveness on machine translation and language modeling, Transformer usually fails on the task with moderate size datasets due to its shortage of inductive bias.

\section{Model}

\tikzstyle{line} = [draw, -latex']
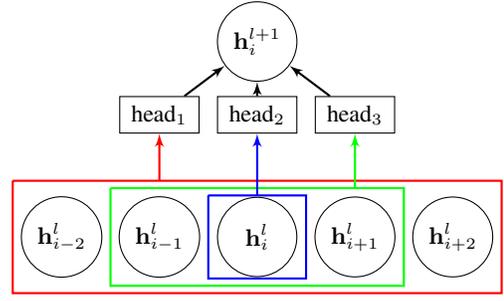
\begin{figure}
\centering
\begin{tikzpicture}[scale=0.65,font=\small\selectfont]
\tikzset{n1/.style={minimum width=3em, draw, circle}}
\tikzset{n2/.style={minimum width=3em, draw, rectangle}}
\node[n2] (B2) at (-2,2.5) {$\text{head}_1$};
\node[n2] (B3) at (0,2.5) {$\text{head}_2$};
\node[n2] (B4) at (2,2.5) {$\text{head}_3$};
\node[n1] (B1) at (0,4) {$\mathbf{h}^{l+1}_i$};
\node[n1] (A1) at (0,0) {$\mathbf{h}^l_i$};
\node[n1] (A2) at (-2,0) {$\mathbf{h}^l_{i-1}$};
\node[n1] (A3) at (-4,0) {$\mathbf{h}^l_{i-2}$};
\node[n1] (A4) at (2,0) {$\mathbf{h}^l_{i+1}$};
\node[n1] (A5) at (4,0) {$\mathbf{h}^l_{i+2}$};
\path[red, draw,thick] (-5,1.15)--(5,1.15)--(5,-1.15)--(-5,-1.15)--(-5,1.15);
\path[green, draw,thick] (-3,1)--(3,1)--(3,-1)--(-3,-1)--(-3,1);
\path[blue, draw,thick] (-1,0.85)--(1,0.85)--(1,-0.85)--(-1,-0.85)--(-1,0.85);
\draw[latex'-,red, draw,thick] (B2)--(-2,1.15);
\draw[latex'-,blue, draw,thick] (B3)--(0,0.85);
\draw[latex'-,green, draw,thick] (B4)--(2,1);
\draw[latex'-,black, draw,thick] (B1)--(B2);
\draw[latex'-,black, draw,thick] (B1)--(B3);
\draw[latex'-,black, draw,thick] (B1)--(B4);
\end{tikzpicture}
\caption{A diagram of Multi-Scale Multi-Head Self-Attention, we can see three heads which correspond to three different scales in the figure. The blue, green, red box illustrate the scale of  $\omega=1$, $\omega=3$, $\omega=5$, respectively. }
    \label{fig:model}
\end{figure}


In the multi-head self-attention, each head captures the pair-wise interactions between words in different feature space. Each head has the same scale of sentence length.

\paragraph{Scale-Aware Self-Attention}
To introduce the concept of multi-scale into the self-attention mechanism, we use a simple way to enhance the regular self-attention, named Scale-Aware Self-Attention (SASA) which is equal to the restricted self-attention which is proposed in \citet{DBLP:conf/nips/VaswaniSPUJGKP17} but using a dynamic size of the window. 

Given a sequence of vectors $\mathbf{H} \in \mathbb{R}^{N \times D}$ with length $N$. SASA has a parameter $\omega$ which is either a constant number or a ratio according to the sequence length to control its working scope. An attention head can be computed as
\begin{align}
    & \text{head}(\mathbf{H}, \omega)_{i,j} = \text{SM}(\frac{\mathbf{Q}_{ij} \text{C}_{ij}(\mathbf{K},\omega)^T}{\sqrt{D}})\text{C}_{ij}(\mathbf{V},\omega), \\
    & \text{C}_{ij}(\mathbf{x},\omega) = [\mathbf{x}_{i,j-\omega},...,\mathbf{x}_{i,j+\omega}], \\
    & \mathbf{Q} = \mathbf{H} \mathbf{W}^Q, \mathbf{K} = \mathbf{H} \mathbf{W}^K, \mathbf{V} = \mathbf{H} \mathbf{W}^V,
\end{align}
where $i$ indicates the i-th head and $j$ means j-th position. The $\text{SM}$ represents ``Softmax" function, $C$ is a function to extract context for a given position. $\mathbf{W}^Q,\mathbf{W}^K,\mathbf{W}^V$ are learnable parameters. The scale of a head could be a variable like $\frac{N}{2}$ or a fixed number like $3$. 

\paragraph{Multi-Scale Multi-Head Self-Attention}
With SASA, we can implement a Multi-Scale Multi-Head Self-Attention (MSMSA) with multiple scale-aware heads. Each head works on different scales. For $N'$ heads self-attention, MSMSA with scales $\Omega=[\omega_1,\cdots,\omega_{N'}]$ is
\begin{align}
     \text{MSMSA}(\mathbf{H},\Omega)& = [\text{head}_1(\mathbf{H},\omega_1);\cdots;\nonumber\\
    & \text{head}_{N'}(\mathbf{H},\omega_{N'})]\mathbf{W}^O,
\end{align}
where $\mathbf{W}^O$ is a parameter matrix.

Compared to the vanilla multi-head self-attention, variable $\Omega$ controls the attended area and makes different heads have different duties.

\paragraph{Multi-Scale Transformer}

With the MSMSA, we could construct the multi-scale Transformer (MS-Transformer. In short, MS-Trans). Besides using MSMSA to replace the vanilla multi-head self-attention, we also remove the FFN (see Eq. \eqref{eq:trans-fnn}) because it could be view as a self-attention with scale $\omega=1$ plus a non-linear activation function. Since MSMSA introduces the locality to the model, the MSMSA with a small scale can be an alternative to the positional embedding. Therefore, the multi-scale Transformer could be formalized as followings.

{\footnotesize
\begin{align}
    \bH_{l+1} = \mathrm{norm}(\bH_l + \text{ReLU}(\mathrm{MSMSA}(\bH_l), \Omega_l))
\end{align}}%
where $l$ is the layer index.

In this work, we limit the choice of scale size among constant numbers $\{1,3,\cdots\}$ or variables depend on the sequence length $\{\frac{N}{16},\frac{N}{8},\cdots \}$. And we force the scale size to be an odd number.

Compared to the hierarchical multi-scale models, multi-scale Transformer allows the attention heads in one layer have various scales of vision, it is a ``soft'' version of viewing different layers as different scales. 

\paragraph{Classification node}
We also find adding a special node at the beginning of each sequence and connecting it directly to the final sentence representation can improve the performance. This technique was introduced in BERT \cite{DBLP:journals/corr/abs-1810-04805}, refer as ``[CLS]". And it is also similar to the ``relay node" in \citet{DBLP:conf/naacl/GuoQLSXZ19}. Different with them, we combine the ``[CLS]" node and the feature from applying max-pooling over all the nodes in the final layer to represent the sentence. There is no difference between the ``[CLS]" node and other tokens in the input sequence except it can directly contribute to the final representation.

\section{Looking for Effective Attention Scales}

Multi-scale Transformer is a flexible module in which each layer can have multi-scale attention heads. Therefore, an important factor is how to design the scale distribution for each layer.

\subsection{Hierarchical Multi-Scale or Flexible Multi-Scale}

A simple way to implement multi-scale feature extraction is following the hierarchical way, which stacks several layers with small-scale heads. The lower layers capture the local features and the higher capture the non-local features.


To verify the ability of hierarchical multi-scale, we design a very simple simulation task named mirrored summation. Given a sequence $\mathbf{A}=\{a_1,...,a_N\}, \quad a \in \mathbb{R}^d$ and drawn from $\mathbf{U}(0,1)$. The target is $\sum_{i=1}^K a_i \odot a_{N-i+1}$, where $K$ is a fixed integer less than the sequence length $N$ and $\odot$ means the Hadamard production. The minimum dependency path length is $N-K$, we use this task to test the ability of models for capturing long-range dependencies. Both train and test set are random generated and they have 200k samples each. We can assume the size of training set is enough to train these models thoroughly.

We use three different settings of MS-Trans.
\begin{enumerate*}
  \item MS-Trans-Hier-S: MS-Transformer with two layers, and each layer has 10 heads with a small scale $\omega=3$.
  \item MS-Trans-deepHier-S: MS-Transformer with six layers, and each layer has 10 heads with a small scale $\omega=3$.
  \item MS-Trans-Flex: MS-Transformer with two layers, and each layer has 10 heads with flexible multi-scales $\omega=3,\frac{N}{16},\frac{N}{8},\frac{N}{4},\frac{N}{2}$. Each scale has two heads.
\end{enumerate*}

\begin{figure}[ht]
    \centering
\resizebox {1.0\linewidth} {!} {
\begin{tikzpicture}[y=.5cm, x=.7cm,font=\footnotesize\sffamily]
	\draw (0,0) -- coordinate (x axis mid) (10,0);
    	\draw (0,0) -- coordinate (y axis mid) (0,10);
    	\foreach \x in {0,1,...,10}
     		\draw (\x,1pt) -- (\x,-3pt)
			node[anchor=north] {\pgfmathparse{int(\x*10)}\pgfmathresult};
    	\foreach \y in {0,1,...,10}
     		\draw (1pt,\y) -- (-3pt,\y)
     			node[anchor=east] {\pgfmathparse{\y*1e-2} \pgfmathprintnumber[fixed,precision=2]\pgfmathresult};
	\node[below=0.8cm] at (x axis mid) {$K$};
	\node[rotate=90, above=0.8cm] at (-1.25, 3) {Test MSE};

	\draw plot[mark=*, mark options={fill=red}, color=red]
		file {toy1a.data};
	\draw plot[mark=triangle*, mark options={fill=blue}, color=blue ]
    	file {toy1e.data};
    \draw plot[mark=triangle*, mark options={fill=red}, color=red ]
    	file {toy1f.data};
    \draw plot[mark=triangle*, mark options={fill=green}, color=green ]
    	file {toy1h.data};
	\draw plot[mark=square*, mark options={fill=black}, color=black]
		file {toy1c.data};

	\begin{scope}[shift={(0.75,5)}]
	\draw (0,0) --
		plot[mark=*, mark options={fill=red}, color=red] (0.25,0) -- (0.5,0)
		node[right]{Trans};
	\draw[yshift=\baselineskip] (0,0) --
		plot[mark=triangle*, mark options={fill=blue}, color=blue] (0.25,0) -- (0.5,0)
		node[right]{MS-Trans-Flex};
	\draw[yshift=2\baselineskip] (0,0) --
		plot[mark=triangle*, mark options={fill=red}, color=blue] (0.25,0) -- (0.5,0)
		node[right]{MS-Trans-Hier-S};
	\draw[yshift=3\baselineskip] (0,0) --
		plot[mark=triangle*, mark options={fill=green}, color=blue] (0.25,0) -- (0.5,0)
		node[right]{MS-Trans-deepHier-S};		
	\draw[yshift=4\baselineskip] (0,0) --
		plot[mark=square*, mark options={fill=black}, color=black] (0.25,0) -- (0.5,0)
		node[right]{BiLSTM};
	\end{scope}
\end{tikzpicture}
}
    \caption{Mirrored Summation Task. The curve of MSE versus valid number $K$ with the sequence length $n=100$ and the dimension of input vectors $d=50$.}
    \label{fig:toy_exp}
\end{figure}
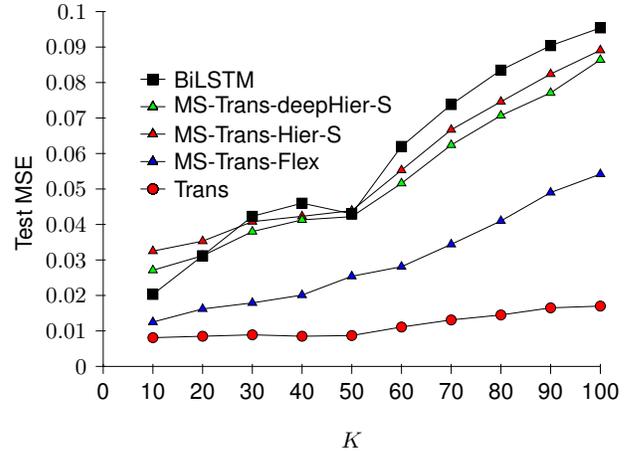

\begin{figure*}[t]
    \centering
\subfloat[]{\label{fig:bert-samelayer}
\begin{tikzpicture}
\begin{axis}[
    width=0.4\linewidth,
    ymin=0, ymax=50,
    minor y tick num = 3,
    area style,
    ylabel={\small{Percentage}},
    ylabel style = {at={(0.05,0.5)}},
    xlabel={Relative Distance of Attention Head},
    xtick={0,5,...,15},
    font=\small,
    ]
\addplot+[ybar interval,mark=no,opacity=0.6] plot coordinates {(0,7.5) (1,7.3) (2,7.4) (3,6.9) (4,6.8) (5,5.9) (6,6.1) (7,6.6) (8,6.5) (9,6.2) (10,5.7) (11,5.6) (12,5.8) (13,5.8) (14,10.0)};
\addlegendentry{head-1 at layer-1}
\addplot+[ybar interval,mark=no,opacity=0.6] plot coordinates {(0,11.5) (1,41.0) (2,21.3) (3,3.0) (4,1.1) (5,0.7) (6, 1.4) (7,1.4) (8,1.1) (9,1.5) (10,1.9) (11,2.1) (12, 2.7) (13, 2.6) (14, 6.8)};
\addlegendentry{head-2 at layer-1}
\addplot+[ybar interval,mark=no,opacity=0.6] plot coordinates {(0,26.1) (1,6.7) (2,5.8) (3, 5.8) (4,5.1) (5,4.5) (6,4.8) (7,4.6) (8,4.2) (9,4.8) (10,4.8) (11,4.6) (12,4.5) (13,4.5) (14,9.2)};
\addlegendentry{head-3 at layer-1}
\end{axis}
\end{tikzpicture}
}
\subfloat[]{\label{fig:bert-difflayer}
\begin{tikzpicture}
\begin{axis}[
width=0.4\linewidth,
    ymin=0, ymax=25,
    minor y tick num = 3,
    area style,
    ylabel={\small{Percentage}},
    ylabel style = {at={(0.05,0.5)}},
    xlabel={\small{Relative Distance of Attention Head}},
    xtick={0,5,...,15},
    font=\small,
    legend style={
    font=\small,
    }
    ]
\addplot+[ybar interval,mark=no,opacity=0.6] plot coordinates {(0,11.1) (1,19.2) (2,10.1) (3,6.9) (4,5.6) (5,4.4) (6,4.4) (7,4.5) (8,4.2) (9,4.2) (10,4.2) (11,4.2) (12,4.6) (13,4.3) (14,8.0)};
\addlegendentry{layer-1}
\addplot+[ybar interval,mark=no,opacity=0.6] plot coordinates {(0,4.8) (1,14.3) (2,9.7) (3,7.7) (4,7.0) (5,6.5) (6,6.2) (7,5.9) (8,5.7) (9,5.3) (10,5.1) (11,4.7) (12,4.6) (13,4.5) (14,8.0)};
\addlegendentry{layer-6}
\addplot+[ybar interval,mark=no,opacity=0.6] plot coordinates {(0,10.1) (1,8.1) (2,7.6) (3,7.3) (4,7.0) (5,6.8) (6,6.6) (7,6.3) (8,6.0) (9,5.7) (10,5.4) (11,5.2) (12,4.9) (13,4.6) (14,8.7)};
\addlegendentry{layer-12}
\end{axis}
\end{tikzpicture}
}
\caption{The visualization of BERT. (a) The attention distance distributions of three heads in the first layer.  The red head only cares about the local pattern and the blue head equally looks at different distances.
(b) The attention distance distribution of different layers. The shallow layer prefers the small scale size and tends to large scale size slowly when the layer gets deeper. Even in the final layer, local patterns still occupy a large percentage.
We truncate the distance at $\frac{N}{2}$ for better visualization. The full figure can be found in the Appendix.}
    \label{fig:bert}
\end{figure*}
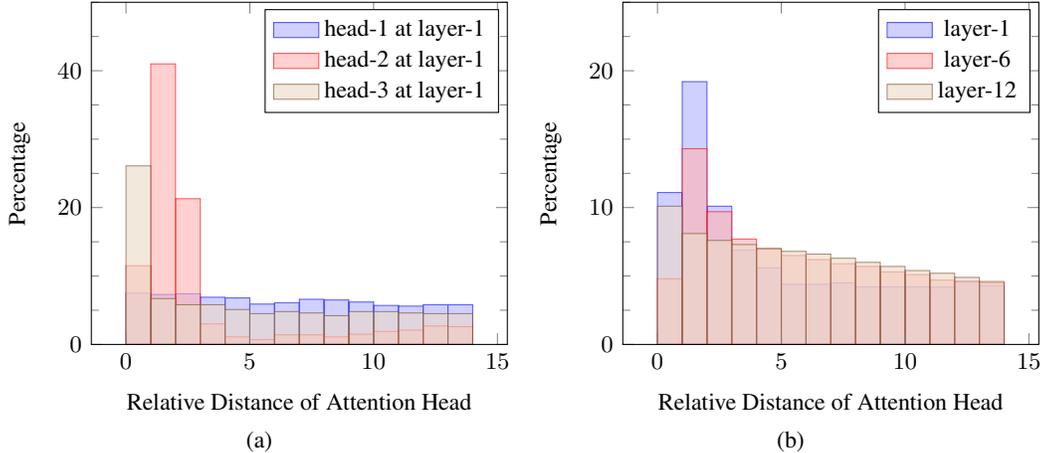

As shown in Fig-\ref{fig:toy_exp}, Transformer achieves the best result, and MS-Trans-Flex follows with it. Although Transformer has the highest potential to capture long-range dependencies, it requires large training samples. Meanwhile, our model balance the data-hungry problem and the ability for capturing long-range dependencies. Based on the comparison of MS-Trans-Hier-S and MS-Trans-deepHier-S, we can find the improvement of additional layers is relatively small.
According to the synthetic experiment and the performance on real tasks (see. Sec-\ref{sec:ana}), we think the large-scale heads are necessary for the lower layers and stacked small-scale heads are hard to capture long-range dependencies. In this case,  a good prior should contain both small-scale and large-scale.

\subsection{Scale Distributions of Different Layers}

Since multi-scale is necessary for each layer, the second question is \textbf{how to design the proportion of different scales for each layer?}
We think each layer may have its preference for the scale distribution. From the linguistic perspective, an intuitive assumption may be: the higher layer has a higher probability for the large-scale heads and the shallow layer has a higher probability for the small-scale heads.


To look for the empirical evidence, we probe several typical cases and analysis the corresponding behavior in a data-driven model, BERT (Bidirectional Encoder Representations from Transformers) \cite{DBLP:journals/corr/abs-1810-04805}.

\paragraph{Analogy Analysis from BERT}
\label{sec:bert}
BERT is a pre-trained Transformer on large scale data, which has shown its power on many NLP tasks and it has good generalization ability. Since BERT is based on Transformer, it is not guided by prior knowledge, their knowledge is learned from data. We probe the behavior of BERT to see whether it fits the linguistic assumption. There are two aspects we want to study, the first is the working scales of attention heads of each layer in BERT. The second is the difference between the scale distributions of different layers, especially the preference of local and global relations. To probe these behaviors, we first run the BERT over many natural sentences and pick the highest activation of the attention map as the attention edge. Then we record the relative distances of these attention edges. In this work, we obtain the data from running BERT on CoNLL03 dataset (see Tab-\ref{tab:exp_all}).

We first draw the Fig-\ref{fig:bert-samelayer} for observing the behavior of heads in the same layer. We pick three heads, and their difference is significant. As shown in the figure, ``head-2'' focus on a certain distance with a small scale, and ``head-1'' cover all the distances. There is a clear division of labor of these heads, and one layer can have both local and global vision via combining features from different heads.

The second figure Fig-\ref{fig:bert-difflayer} shows the trend of distance preference when the depth of layer increased. We can find the model move from local vision to global vision slowly and shallow layers have a strong interest in local patterns.

The visualization of BERT fits the design of Multi-Scale Transformer, the first observation corresponds to the design of multi-scale multi-head self-attention which ask different heads focus on different scales, and the second observation provides a good reference of the trend of scale distribution across layers. Using such knowledge can largely reduce the requirement of training data for Transformer-like models.

\subsection{Control Factor of Scale Distributions for Different Layers}

From the intuitive linguistic perspective and empirical evidence, we design a control factor of scale distributions for different layers of multi-scale Transformer.

Let $L$ denote the number of layers in multi-scale Transformer, $|\Omega|$ denote the number of candidate scale sizes, and $n^l_k$ denote the number of heads for $l$-th layer and $k$-th scale size.

The head number $n^l_k$ is computed by
\begin{align}
    z^l_k &= \begin{cases}
        0 & l=L  ~~\text{or}~~  k= |\Omega| \\
        z^l_{k+1} + \frac{\alpha}{l} & k\in \{0,\cdots,|\Omega|-1\} \\
    \end{cases} \label{eq:scale} \\
    n^l &= \text{softmax}(z^l) \cdot N'
\end{align}

In the above equations, we introduce a hyper-parameter $\alpha$ to control the change of preference of scale sizes for each layer. For example, $\alpha=0$ means all the layers use the same strategy of scale size, $\alpha>0$ means the preference of smaller scale increased with the decline of layer depth, and $\alpha<0$ indicates the preference of larger scale increased with the decline of layer depth. As the conclusion of analyzing BERT, we believe the deep layer has a balanced vision over both local and global patterns, so the top layer should be set to looking all the scale size uniformly. More specific, when $\alpha=0.5$ and $N'=10$, three layers have $n^{l=1}=\{5,2,2,1,0\},n^{l=2}=\{4,2,2,1,1\},n^{l=3}=\{2,2,2,2,2\}$, it represents the first layer has 5 head with scale size of 1, 2 head with scale size of 3, 2 head with scale size of $\frac{N}{16}$ and 1 head with scale size of $\frac{N}{8}$.

\begin{table*}[t]
    \centering\small
    \caption{An overall of datasets and its hyper-parameters, ``H DIM, $\alpha$, head DIM" indicates the dimension of hidden states, the hyper-parameter for controlling the scale distribution, the dimension of each head, respectively. The candidate scales are $1,3,\frac{N}{16},\frac{N}{8},\frac{N}{4}$ for SST,MTL-16,SNLI datasets. And we use $1,3,5,7,9$ for sequence labeling tasks. MTL-16$^\dagger$ consists of 16 datasets, each of them has 1400/200/400 samples in train/dev/test.}
    \label{tab:exp_all}
    \begin{tabular}{c|c*7{c}}
    \toprule
    \multicolumn{2}{c}{Dataset} & Train & Dev. & Test & $|V|$ & H DIM & $\alpha$ & head DIM \\
    \midrule
    \multicolumn{2}{c}{SST \cite{DBLP:conf/emnlp/SocherPWCMNP13}} & 8k &1k& 2k & 20k & 300 & 0.5 & 30\\
    \midrule
    \multirow{5}{1.5cm}{MTL-16 $^\dagger$ \cite{DBLP:conf/acl/LiuQH17}} & \multirow{5}{4cm}{{\small Apparel Baby Books Camera DVD Electronics Health IMDB Kitchen Magazines MR Music Software Sports Toys Video} } & \multirow{5}{*}{1400} & \multirow{5}{*}{200} & \multirow{5}{*}{400} & \multirow{5}{*}{8k$\sim$28k} & \multirow{5}{*}{300} & \multirow{5}{*}{0.5} & \multirow{5}{*}{30} \\
    & & & & & & & &\\
    & & & & & & & &\\
    & & & & & & & &\\
    & & & & & & & &\\
    \midrule
    \multicolumn{2}{c}{PTB POS \cite{DBLP:journals/coling/MarcusSM94}} & 38k &5k& 5k & 44k & 300 & 1.0 & 30 \\
    \midrule
    \multicolumn{2}{c}{CoNLL03 \cite{DBLP:conf/conll/SangM03}} &15k & 3k& 3k & 25k & 300 & 1.0 & 30 \\
    \midrule
    \multicolumn{2}{c}{CoNLL2012 NER \cite{DBLP:conf/conll/PradhanMXUZ12}} &94k &14k& 8k & 63k & 300 & 1.0 & 30 \\
    \midrule
    \multicolumn{2}{c}{SNLI \cite{DBLP:conf/emnlp/BowmanAPM15}} & 550k &10k & 10k & 34k & 600 & 0.5 & 64 \\

    \bottomrule
    \end{tabular}
\end{table*}

\section{Experiments}
\label{sec:exp}

We evaluate our model on 17 text classification datasets, 3 sequence labeling datasets and 1 natural language inference dataset. All the statistics can be found in Tab-\ref{tab:exp_all}. Besides, we use GloVe \cite{pennington2014glove} to initialize the word embedding and JMT \cite{DBLP:conf/emnlp/HashimotoXTS17} for character-level features. The optimizer is Adam \cite{DBLP:journals/corr/KingmaB14} and the learning rate and dropout ratio are listed in the Appendix. 

To focus on the comparison between different model designs, we don't list results of BERT-like models because the data augmentation and pre-training is an orthogonal direction.
 \begin{table}[t]
    \centering\small
    \caption{Test Accuracy on SST dataset.}
    \label{tab:exp_cls1}
    \begin{tabular}{lc}
        \toprule
        Model & Acc \\
        \midrule
        BiLSTM \cite{DBLP:conf/emnlp/LiLJH15} & 49.8  \\

        Tree-LSTM \cite{DBLP:conf/acl/TaiSM15}  & 51.0 \\

        CNN-Tensor \cite{DBLP:conf/emnlp/LeiBJ15}  & 51.2 \\

        Emb + self-att \cite{DBLP:conf/aaai/ShenZLJPZ18}  & 48.9 \\

        BiLSTM + self-att \cite{DBLP:journals/corr/abs-1808-07383}  & 50.4 \\

        CNN + self-att \cite{DBLP:journals/corr/abs-1808-07383} & 50.6  \\
        Dynamic self-att \cite{DBLP:journals/corr/abs-1808-07383}  & 50.6 \\

        DiSAN \cite{DBLP:conf/aaai/ShenZLJPZ18}& 51.7  \\

        \midrule
        Transformer & 50.4 \\

        Multi-Scale Transformer &  \textbf{51.9}   \\
        \bottomrule
    \end{tabular}

\end{table}

\begin{table}[t]
    \centering\small\setlength{\tabcolsep}{3pt}
    \caption{Test Accuracy over MTL-16 datasets. ``SLSTM" refer to the sentence-state LSTM \cite{DBLP:conf/acl/ZhangLS18}. }
    \label{tab:exp_cls2}    
    \begin{tabular}{l*4{c}}
        \toprule
        \bf \multirow{2}{*}{Dataset} & \multicolumn{4}{c}{Acc (\%)}\\
        \cmidrule(lr){2-5}
        & MS-Trans & Transformer & BiLSTM & SLSTM  \\
        \midrule
        Apparel & 87.25 & 82.25 & 86.05 & 85.75  \\

        Baby & 85.50 & 84.50 & 84.51 & 86.25  \\

        Books & 85.25  & 81.50 & 82.12 & 83.44  \\

        Camera &  89.00 & 86.00 & 87.05 & 90.02  \\

        DVD & 86.25 & 77.75 & 83.71 & 85.52  \\

        Electronics & 86.50 & 81.50 & 82.51 & 83.25  \\

        Health & 87.50 & 83.50 & 85.52 & 86.50  \\

        IMDB & 84.25 & 82.50 & 86.02 & 87.15  \\

        Kitchen & 85.50 & 83.00 & 82.22 & 84.54  \\

        Magazines & 91.50  & 89.50 & 92.52 & 93.75  \\

        MR & 79.25 & 77.25 & 75.73 & 76.20 \\

        Music &  82.00& 79.00 & 78.74 & 82.04 \\

        Software & 88.50& 85.25 & 86.73 & 87.75  \\

        Sports & 85.75& 84.75 & 84.04 & 85.75  \\

        Toys & 87.50& 82.00 &  85.72 & 85.25  \\

        Video & 90.00 & 84.25 & 84.73 & 86.75  \\

        \midrule
        \bf Average & \textbf{86.34} & 82.78 & 84.01 & 85.38 \\
        \bottomrule
    \end{tabular}
\end{table}

\subsection{Text Classification}

Text Classification experiments are conducted on Stanford Sentiment Treebank(SST) dataset \cite{DBLP:conf/emnlp/SocherPWCMNP13} and MTL-16 \cite{DBLP:conf/acl/LiuQH17} consists of 16 small datasets in different domains. Besides the base model we introduced before, we use a two-layer MLP(Multi-Layer Perceptron) with softmax function as the classifier. It receives the feature from applying max-pooling over the top layer plus the classification node.  

Tab-\ref{tab:exp_cls1} and \ref{tab:exp_cls2} give the results on SST and MTL-16. Multi-Scale Transformer achieves 1.5 and 3.56 points against Transformer on these two datasets, respectively. Meanwhile, Multi-Scale Transformer also beats many existing models including CNNs and RNNs.

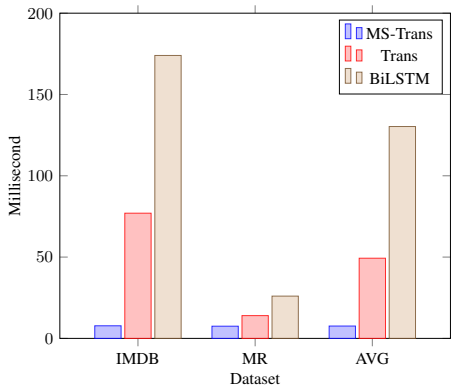
\begin{figure}[t]
    \centering
\begin{tikzpicture}[scale=0.7]
\begin{axis}[
    ymin=0, ymax=200,
    xmin=0, xmax=10,
    ylabel={\small{Millisecond}},
    ylabel style = {at={(0.05,0.5)}},
    xlabel={Dataset},
    ybar,
    width=9cm,
    bar width=0.5cm,
    xticklabels={IMDB,MR,AVG},
    xtick=data,
    font=\small,
    ]
\addplot+[mark=no,opacity=0.8] plot coordinates {(2,7.75) (5,7.5) (8,7.60)};
\addlegendentry{MS-Trans}
\addplot+[mark=no,opacity=0.8] plot coordinates {(2,77) (5, 14) (8,49.31)};
\addlegendentry{Trans}
\addplot+[mark=no,opacity=0.8] plot coordinates {(2,174) (5, 26) (8,130.3)};
\addlegendentry{BiLSTM}
\end{axis}
\end{tikzpicture}
    \caption{Test time per batch (batch size is 128) on the dataset which has the longest length (IMDB), the dataset which has the shortest length (MR), and the  average over 16 datasets in MTL-16.} 
    \label{fig:speed}
\end{figure}

Since the sentence average length of MTL-16 dataset is relatively large, we also report the efficiency result in Fig-\ref{fig:speed}. We implement the MS-Trans with Pytorch\footnote{https://pytorch.org} and DGL\cite{DBLP:journals/corr/abs-1909-01315}. Multi-Scale Transformer achieves 6.5 times acceleration against Transformer on MTL-16 dataset on average (average sentence length equals 109 tokens). The maximum of acceleration reaches 10 times (average 201 tokens) and Multi-Scale Transformer can achieve 1.8 times acceleration on very short sentences (average 22 tokens). 

\subsection{Sequence Labelling}
\begin{table*}[t]
    \centering\small
    \caption{Results on sequence labeling tasks. We list ``Advanced Techniques'' except pre-trained embeddings (GloVe, Word2Vec, JMT) in columns. The ``Char" indicates character-level features, it also includes the Capitalization Features, Lexicon Features, etc. The ``CRF" means an additional Conditional Random Field layer.}
    \label{tab:seq_exp}
    \begin{tabular}{p{6cm}|*2{c}|c|*2{c}}
    \toprule
        \multirow{3}{*}{Model} & \multicolumn{2}{|c|}{\multirow{2}{*}{Adv Tech}} & POS & \multicolumn{2}{|c}{NER} \\
        \cline{4-6}
        & \multicolumn{2}{|c|}{} &  PTB  & CoNLL2003 & CoNLL2012 \\
        \cline{2-6}
        &char & CRF  &  Acc & F1 & F1 \\
        \midrule
        \citet{DBLP:conf/emnlp/LingDBTFAML15} &\checkmark & \checkmark  & 97.78  & - & - \\


        \citet{DBLP:journals/corr/HuangXY15} &\checkmark & \checkmark  & 97.55 & 90.10 & - \\

        \citet{chiu2016sequential} &\checkmark &\checkmark   &  -  & 90.69 & 86.35 \\

        \citet{DBLP:conf/acl/MaH16} &\checkmark & \checkmark  & 97.55 &  91.06 & - \\


        \citet{DBLP:journals/tacl/ChiuN16} &\checkmark & \checkmark & - & 91.62 & 86.28 \\

        \citet{DBLP:conf/acl/ZhangLS18}& \checkmark & \checkmark  & 97.55  & 91.57 & - \\

        \citet{DBLP:journals/corr/abs-1808-03926} &\checkmark & \checkmark & 97.43  & 91.11 & 87.84 \\

        \midrule
        Transformer& & &  96.31  & 86.48 & 83.57 \\

        Transformer + Char& \checkmark & &  97.04  & 88.26 & 85.14 \\

        Multi-Scale Transformer& & & 97.24 & 89.38 & 85.26 \\

        Multi-Scale Transformer + Char& \checkmark & &  97.54 & 91.33  & 86.77 \\

        Multi-Scale Transformer + Char + CRF& \checkmark & \checkmark &  \textbf{97.66} & \textbf{91.59}  & \textbf{87.80}  \\
        \bottomrule
    \end{tabular}
\end{table*}

Besides tasks which use the model as a sentence encoder, we are also interested in the effectiveness of our model on sequence labeling tasks. We choose the Part-of-Speech (POS) tagging and the Named Entity Recognition (NER) task to verify our model. We use three datasets as our benchmark: Penn Treebank (PTB) POS tagging dataset \cite{DBLP:journals/coling/MarcusSM94}, CoNLL2003 NER dataset \cite{DBLP:conf/conll/SangM03}, CoNLL2012 NER dataset \cite{DBLP:conf/conll/PradhanMXUZ12}.

Results in Tab-\ref{tab:seq_exp} shows Multi-Scale Transformer beats the vanilla Transformer on these three sequence labeling datasets, which consists of other results reported above. It shows Multi-Scale Transformer can extract useful features for each position as well.

\subsection{Natural Language Inference}

\begin{table}[t]
    \centering\small
    \caption{Test Accuracy on SNLI dataset for sentence vector-based models. }
    \label{tab:nli_exp}
    \begin{tabular}{p{6.5cm}c}
    \toprule
        Model & Acc\\
    \midrule
        BiLSTM \cite{DBLP:journals/corr/LiuSLW16} & 83.3  \\
        BiLSTM + self-att \cite{DBLP:journals/corr/LiuSLW16} & 84.2   \\



        4096D BiLSTM-max \cite{DBLP:conf/emnlp/ConneauKSBB17} & 84.5   \\
        300D DiSAN \cite{DBLP:conf/aaai/ShenZLJPZ18} & 85.6  \\
        Residual encoders \cite{DBLP:conf/repeval/NieB17} & 86.0 \\
        Gumbel TreeLSTM \cite{DBLP:conf/aaai/ChoiYL18} & 86.0  \\

        Reinforced self-att \cite{DBLP:conf/ijcai/ShenZLJWZ18} & 86.3 \\
        2400D Multiple DSA \cite{DBLP:journals/corr/abs-1808-07383} & 87.4  \\
        \midrule

        Transformer & 82.2 \\

        Multi-Scale Transformer & \textbf{85.9} \\
    \bottomrule
    \end{tabular}
\end{table}

Natural Language Inference (NLI) is a classification which ask the model to judge the relationship of two sentences from three candidates, ``entailment", ``contradiction", and ``neutral". We use a widely-used benchmark Stanford Natural Language Inference (SNLI) \cite{DBLP:conf/emnlp/BowmanAPM15} dataset to probe the ability of our model for encoding sentence, we compare our model with sentence vector-based models. Different with the classifier in text classification task, we follow the previous work \cite{DBLP:conf/acl/BowmanGRGMP16} to use a two-layer MLP classifier which takes $\text{concat}( \mathbf{r}_1, \mathbf{r}_2, \| \mathbf{r}_1 - \mathbf{r}_2 \|, \mathbf{r}_1 - \mathbf{r}_2) $ as inputs, where $ \mathbf{r}_1, \mathbf{r}_2$ are representations of two sentences and equals the feature used in text classification task.

As shown in Tab-\ref{tab:nli_exp}, Multi-Scale Transformer outperforms Transformer and most classical models, and the result is comparable with the state-of-the-art. The reported number of Transformer is obtained with heuristic hyper-parameter selection, we use a three-layer Transformer with heavy dropout and weight decay. And there is still a large margin compared to Multi-Scale Transformer. This comparison also indicates the moderate size training data (SNLI has 550k training samples) cannot replace the usefulness of prior knowledge.

\begin{table}[t]\small
    \centering
    \caption{Analysis of different scale distributions on SNLI test set. The upper part shows the influence of hyper-parameter $\alpha$ which change the distribution of scales across layers. The five candidates of scale size are $1,3,\frac{N}{16},\frac{N}{8},\frac{N}{4}$, respectively. The lower part lists the performance of single-scale models which use a fixed scale for the whole model.}
    \label{tab:scale_exp}
    \begin{tabular}{c|c|c|c|c}

             \toprule
          \textbf{multi-scale} & $\alpha$& N' & L & Acc \\
\midrule
         A & 1.0 & 5 & 3 & 85.5 \\
         B & 0.5 & 5 & 3 & \textbf{85.9}\\
         C & 0.0 & 5 & 3 & 84.9\\
         D & -0.5 & 5 & 3 & 84.7\\
         E & -1.0 & 5 & 3 & 84.3\\

         \midrule
         \textbf{single-scale} & \multicolumn{2}{c|}{$\omega$}& L & Acc \\
         \midrule
         F & \multicolumn{2}{c|}{$3$} & 3 & 84.3 \\
         G & \multicolumn{2}{c|}{$N/16$} & 3 & 83.9  \\
         H & \multicolumn{2}{c|}{$N/8$} & 3 & 81.7 \\
         I & \multicolumn{2}{c|}{$N/4$} & 3 & 80.7 \\
         \bottomrule
    \end{tabular}
\end{table}

\subsection{Analysis}
\label{sec:ana}

\subsubsection{Influence of Scale Distributions}

As we introduced in Eq. \eqref{eq:scale}, we control the scale distributions over layers by a hyper-parameter $\alpha$. In this section, we give a comparison of using different $\alpha$, where the positive value means local bias increased with the decline of layer depth and the negative value means global bias increased with the decline of layer depth.

As shown in the upper part of Tab-\ref{tab:scale_exp}, local bias in shallow layers is a key factor to achieve good performance, and an appropriate positive $\alpha$ achieves the best result. In contrast, all the negative values harm the performance, that means too much global bias in shallow layers may lead the model to a wrong direction. The observation of this experiment fits our intuition, the high-level feature is the composition of low-level terms.

\subsubsection{Multi-Scale vs. Single-Scale}

As we claimed before, Multi-Scale Transformer can capture knowledge at different scales at each layer. Therefore, a simple question needs to be evaluated is whether the multi-scale model outperforms the single-scale model or not. To answer this question, we compare Multi-Scale Transformer with several single-scale models. Model F,G,H,I have the same number of layers and attention heads with Multi-scale Transformer, but their scales are fixed. 

Result in the lower part of Tab-\ref{tab:scale_exp} reveal the value of Multi-Scale Transformer, it achieves 1.6 points improvement against the best single-scale model. And this result also supports that local bias is an important inductive bias for NLP task.

\section{Related Works}

\paragraph{Typical multi-scale models}
The multi-scale structure has been used in many NLP models, it could be implemented in many different ways. Such as stacked layers \cite{kalchbrenner2014convolutional,kim2014convolutional}, tree-structure \cite{DBLP:conf/emnlp/SocherPWCMNP13,DBLP:conf/acl/TaiSM15,zhu2015long}, hierarchical timescale \cite{el1995hierarchical,chung2016hierarchical}, layer-wise gating \cite{DBLP:conf/icml/ChungGCB15}. Since these models are built on modules like RNNs and CNNs, which embodies the intrinsic local bias by design, the common spirit of introducing multi-scale is to enable long-range communications. In contrast, Transformer allows long-range communications, so we want the multi-scale brings local bias.

\paragraph{Transformers with additional inductive bias}
This work is not the first attempt of introducing inductive bias into Transformer.

\citet{DBLP:conf/naacl/ShawUV18} suggest Transformer should care about the relative distance between tokens rather than the absolute position in the sequence. The information of relative distance could be obtained by looking multi-scale of the same position, so our model could be aware of the relative distance if using enough scales.

\citet{DBLP:conf/emnlp/LiTYLZ18} propose a regularization of enhancing the diversity of attention heads. Our multi-scale multi-head self-attention can make a good division of labor of heads via restricting them in different scales.

\citet{DBLP:conf/emnlp/YangTWMCZ18} and \citet{DBLP:journals/corr/abs-1810-13320} also introduce the local bias into Transformer.

Different from the above models, we focus on importing the notion of multi-scale to self-attention. 
Meanwhile, their models use the single-scale structure. Our experimental results have shown the effectiveness of the multi-scale mechanism.

\section{Conclusion}
In this work, we present Multi-Scale Self-Attention and  Multi-Scale Transformer which combines the prior knowledge of multi-scale and the self-attention mechanism. As a result, it has the ability to extract rich and robust features from different scales. We compare our model with the vanilla Transformer on three real tasks (21 datasets). The result suggests our proposal outperforms the vanilla Transformer consistently and achieves comparable results with state-of-the-art models. 

\section{Acknowledgments}
This work was supported by the National Key Research and Development Program of China (No. 2018YFC0831103), National Natural Science Foundation of China (No. 61672162), Shanghai Municipal Science and Technology Major Project (No. 2018SHZDZX01) and ZJLab.

\fontsize{9.0pt}{10.0pt}\selectfont 
\bibliography{aaai2020}
\bibliographystyle{aaai}

\end{document}